\documentclass{article}

\PassOptionsToPackage{numbers, compress}{natbib}

\usepackage[preprint]{nips_2018}

\usepackage[utf8]{inputenc} 
\usepackage[T1]{fontenc}    
\usepackage{hyperref}       
\usepackage{url}            
\usepackage{booktabs}       
\usepackage{amsfonts}       
\usepackage{nicefrac}       
\usepackage{microtype}      

\usepackage{amsmath}
\usepackage{amssymb}
\usepackage{graphicx}
\usepackage{caption}
\usepackage{xspace}

\newcommand{\tgan}{TGAN\xspace}
\newcommand{\sdv}{GC\xspace}
\newcommand{\dsid}{BN-Id\xspace}
\newcommand{\dsco}{BN-Co\xspace}

\newcommand{\R}{\mathbb{R}}
\newcommand{\pr}{\mathbb{P}}
\newcommand{\E}{\mathbb{E}}

\usepackage{xcolor}
\usepackage{soul}
\definecolor{myhlcolor}{HTML}{DDDDDD}
\sethlcolor{myhlcolor}

\usepackage{tikz}
\usepackage{pgfplots}
\pgfplotsset{compat=1.8}
\usepackage{pgfplotstable}
\usepackage{sansmath}
\pgfplotsset{compat=1.13}
\pgfplotsset{
/pgfplots/colormap={blueblue}{rgb255=(255,255,255) rgb255=(0,0,127)}
}
\usepackage{adjustbox}
\usepackage{xcolor}

\DeclareMathOperator*{\argmax}{arg\,max}

\title{Synthesizing Tabular Data\\using Generative Adversarial Networks}

\author{
  Lei Xu \\
  LIDS, MIT\\
  Cambridge, MA\\
  \texttt{leix@mit.edu} \\
 \And
  Kalyan Veeramachaneni \\
  LIDS, MIT\\
  Cambridge, MA\\
  \texttt{kalyanv@mit.edu} \\
}

\begin{document}

\maketitle

\begin{abstract}
Generative adversarial networks (GANs) implicitly learn the probability distribution of a dataset and can draw samples from the distribution. This paper presents, Tabular GAN (\tgan), a generative adversarial network which can generate tabular data like medical or educational records. Using the power of deep neural networks, \tgan generates high-quality and fully synthetic tables while simultaneously generating discrete and continuous variables. When we evaluate our model on three datasets, we find that \tgan outperforms conventional statistical generative models in both capturing the correlation between columns and scaling up for large datasets. 

\end{abstract}

\section{Introduction}
Today, organizations are increasingly using machine learning on relational tabular data to intelligently augment processes and workflows usually carried out by humans. According to a recent survey performed by the data science platform \textsc{Kaggle}, tabular data is the most commonly encountered data type in business, and the second most common format in academia \cite{kaggle2017state}. At the same time, researchers are touting synthetic data for its ability to alleviate a number of common data science concerns, including resolving critical bottlenecks \cite{che2017boosting,patki2016synthetic}, clearing bureaucratic hurdles encountered in data access, and providing a ``safe data space'' for exploration \cite{choi2017generating, patel2018correlated}. Synthetic datasets can be generated to fit specific needs, like testing new tools or creating educational tutorials using them or can eliminate unwanted contingencies when sharing data: for example, a company might give its employees synthetic data to prevent them from having access to data that may pertain to their friends or to celebrities, or it might provide synthetic data to external consultants to eliminate risk in the event of an accidental breach. 

During the past decade, synthetic data generation has been accomplished by modeling a joint multivariate probability distribution for a given dataset $\pr(\mathbf{D})$ and then sampling from that distribution. Complicated datasets have required more complex distributions: for example, a sequence of events may have been modeled using hidden Markov models, or a set of non-linearly correlated variables could be modeled using \textit{copulas}. Nevertheless, these generative models are restricted by the type of distribution functions available to users, severely limiting the representations that can be used to create generative models and subsequently limiting the fidelity of the synthetic data.  

At the same time, some researchers in the statistical sciences community have begun been using randomization-based methods to generate synthetic data \cite{reiter2005releasing, raghunathan2003multiple,reiter2004simultaneous,kinney2011towards}. Most of these efforts aimed at enabling data disclosure, simultaneously imputing and disclosing data, and preserving the privacy of the people represented by the data (generally, respondents of a survey). The creation of generative models using neural models like variational auto-encoders \cite{kingma2013auto}, and, subsequently, generative adversarial networks (GAN) \cite{goodfellow2014generative} and their numerous extensions, is appealing in terms of both the performance and flexibility offered in representing data and the promise of generating and manipulating images and natural languages. 

In this paper, we develop a synthetic data generator based on generative adversarial networks for tabular data. We focus on generating tabular data with mixed variable types (multinomial/discrete and continuous) and propose \tgan. To achieve this, we use LSTM with attention in order to generate data column by column. To asses, we first statistically evaluate the synthetic data generated by \tgan. We also demonstrate that machine learning models trained on data generated using \tgan can achieve significantly higher performance than models trained on data generated from other competitive data synthesizers that rely on multivariate probabilistic graphical models. 

The paper is organized as follows: section~\ref{related} presents a brief synopsis of related work in GANs and previous synthetic data synthesizers; section~\ref{GANmodel} introduces our TGAN model; section~\ref{seceva} and section~\ref{secres} explain experiment settings and results; and section~\ref{seccon} gives our conclusion. 

\section{Related Work}\label{related}
\noindent \textbf{Generative Adversarial Networks} Since GANs were first proposed, many efforts have been made to speed up and stabilize the training process \cite{salimans2016improved, arjovsky2017wasserstein, gulrajani2017improved, berthelot2017began}, with application-based studies mainly focusing on generating images. GANs can generate high-quality images \cite{radford2015unsupervised, denton2015deep, karras2017progressive}, and some models can even generate images conditioned on images or texts \cite{mirza2014conditional, reed2016generative, isola2017image, zhu2017unpaired, kim2017learning, yi2017dualgan}. However, generating discrete variables is a challenge for GANs: the authors in \cite{kusner2016gans, che2017maximum} attempt a differential model by designing special functions or modifying the loss function; other researchers \cite{yu2017seqgan, fedus2018maskgan, xu2018dp} use a reinforcement learning framework to train the non-differentiable model, making natural language generation possible. Other GAN applications include information retrieval \cite{wang2017irgan}, dialogue systems \cite{li2017adversarial}, and speech processing \cite{pascual2017segan}. 

\noindent \textbf{Synthetic Data Generation}
Synthetic data is useful in data science, as shown in \citet{howe2017synthetic}'s thorough analysis of such data's use cases and social benefits. There are several statistical ways to generate this kind of synthetic data, including classification and regression trees \cite{reiter2005using, nowok2016synthpop} and Bayesian networks \cite{dong2014nonparametric, oliva2016bayesian, valera2017general}. \citet{ping2017datasynthesizer} introduce a web-based Data Synthesizer using a Bayesian network to model the correlation between features. \citet{patki2016synthetic} propose a framework to recursively generate a relational database and use \textit{copulas}. Besides statistical models generating fully synthetic data, neural models are used to impute missing values in datasets; for example, \citet{gondara2018mida} uses deep de-noising autoencoders, and \citet{yoon2018gain} use GAN. Recently, several GAN models emerged to handle tabular data, especially to generate medical records. \texttt{RGAN} and \texttt{RCGAN} \cite{esteban2017real} can generate real-valued time-series data. \texttt{medGAN} \cite{choi2017generating}, \texttt{corrGAN} \cite{patel2018correlated} and several improved models \cite{avino2018generating, camino2018generating,yahi2017generative, beaulieu2017privacy,walonoski2017synthea} can generate discrete medical records but do not tackle the complexity in generating multimodal continuous variables. \texttt{ehrGAN} \cite{che2017boosting} generates augmented medical records but doesn't explicitly generate synthetic data. 

Finally, perhaps the work that is closest to our work is \texttt{tableGAN} \cite{park2018data} - that is, it tries to solve the problem of generating synthetic data for a tabular dataset\footnote{\texttt{tableGAN} was released in June 2018 and we were informed about its existence during a review process for this paper - submitted in May 2018.}. However, there are a few fundamental differences. It uses convolutional neural networks while we use recurrent networks. Also, \texttt{tableGAN}  explicitly optimizes the prediction accuracy on synthetic data by minimizing cross entropy loss while our model cares more about marginal distribution. We explicitly learn the marginal distribution of each column by minimizing KL divergence. 

\section{GANs for tabular data}\label{GANmodel}
Developing a general-purpose GAN that would reliably work for a tabular dataset is nontrivial. Complexities arise due to the various types of data that can be present in the table, including numerical, categorical, time, text, and cross-table references. This is in addition to the variety of shapes the distributions of these variables can take, including multimodal, long tail, and several others. We begin by formalizing the synthetic table generation task and describing the mechanisms that evaluate how well synthetic data actually achieve the goals described in the previous section. 

\noindent \textbf{Synthetic table generation task}: A table $\mathbf{T}$ contains $n_c$ continuous random variables - $\{C_1, \ldots, C_{n_c}\}$, and $n_d$ discrete (multinomial) random variables $\{D_1, \ldots, D_{n_d}\}$. These variables follow an unknown joint distribution $\pr(C_{1:n_c}, D_{1:n_d})$. Each row is one sample from the joint distribution represented using a lowercase $\{c_{1,j}, \ldots, c_{n_c,j}, d_{1, j}, \ldots, d_{n_d, j}\}$. Each row is sampled independently; that is, we do not consider sequential data. The goal is to learn a generative model $\mathtt{M}(C_{1:n_c}, D_{1:n_d})$ such that samples generated from this model $\mathtt{M}$ create a synthetic table $\mathbf{T}_{synth}$ that can satisfy the following requirements. (1) A machine learning model learned using $\mathbf{T}_{synth}$ can achieve a similar accuracy on a real test table $\mathbf{T}_{test}$ (usually set aside at the beginning), as would a model learned using the data from table $\mathbf{T}$. (2) Mutual information: The mutual information between an arbitrary pair of variables $i,j$ in $\mathbf{T}$ and $\mathbf{T_{synth}}$ is similar.

\subsection{Reversible Data Transformation}
To enable neural networks to learn the model effectively we apply a series of reversible transformations to the variables in the table. Neural networks can effectively generate values with a distribution centered over $(-1, 1)$ using $\tanh$, as well as a low-cardinality multinomial distribution using softmax. Thus, we convert a numerical variable into a scalar in the range $(-1, 1)$ and a multinomial distribution, and convert a discrete variable into a multinomial distribution.

\noindent \textbf{Mode-specific normalization for numerical variables}: Numerical variables in tabular datasets sometimes follow a multimodal distribution. We use a Gaussian kernel density estimation to estimate the number of modes of a continuous variable. In the three datasets we use this paper, we found that $4/7$ variables in the Census dataset, $22/27$ continuous variables in the KDD99 dataset, and $1/10$ variables in the Covertype dataset have multiple modes. Simply normalizing numerical feature to $[-1, 1]$ and using $\tanh$ activation to generate these features does not work well. For example, if there is a mode close to $-1$ or $1$, the gradient will saturate when back-propagating through $\tanh$. 

To effectively sample values from a multimodal distribution, we cluster values of a numerical variable using a Gaussian Mixture model (GMM). 
\begin{itemize}
\item We train a GMM with $m$ components for each numerical variable $C_i$. GMM models a distribution with a weighted sum of $m$ Gaussian distributions. The means and standard deviations of the $m$ Gaussian distributions are $\eta_i^{(1)}, \ldots, \eta_i^{(m)}$ and $\sigma_i^{(1)}, \ldots, \sigma_i^{(m)}$. 
\item We compute the probability of $c_{i, j}$ coming from each of the $m$ Gaussian distributions as a vector $u_{i,j}^{(1)}, \ldots u_{i,j}^{(m)}$. $u_{i, j}$ is a normalized probability distribution over $m$ Gaussian distributions. 
\item We normalize $c_{i, j}$ as $v_{i, j} = (c_{i, j} - \eta_i^{(k)}) / 2\sigma_i^{(k)}$, where $k=\argmax_k u_{i, j}^{(k)}$. We then clip $v_{i,j}$ to $[-0.99, 0.99]$. 
\end{itemize}
Then we use $u_i$ and $v_i$ to represent $c_i$. For simplicity, we cluster all the numerical features, i.e. both uni-modal and multi-modal features are clustered to $m=5$ Gaussian distributions. The simplification is fair because GMM automatically weighs $m$ components. For example, if a variable has only one mode and fits some Gaussian distribution, then GMM will assign a very low probability to $m-1$ components and only $1$ remaining component actually works, which is equivalent to not clustering this feature.

\noindent \textbf{Smoothing for categorical variables}: In generating categorical variables the model faces a similar challenge it faces in natural language generation, which is how to make the model differentiable. In natural language generation, people use reinforcement learning \cite{yu2017seqgan} or Gumbel softmax \cite{kusner2016gans} to deal with this issue. We are facing a similar challenge but the number of categories is much smaller than the size of vocabulary in the natural language. So we can generate the probability distribution directly using softmax. But we find it necessary to convert categorical variables to one-hot-encoding representation and add noise to binary variables.
\begin{itemize}
    \item A sample $d_{i, j}$ of a discrete variable $D_i$ is first represented as a $|D_i|$-dimensional one-hot vector $\mathbf{d}_{i, j}$.
    \item We then add noise to each dimension as $\mathbf{d}_{i,j}^{(k)}\leftarrow \mathbf{d}_{i, j}^{(k)} + \mathtt{Uniform}(0, \gamma)$. We set $\gamma=0.2$. 
    \item We then renormalize the representation as $\mathbf{d}_{i,j}\leftarrow \mathbf{d}_{i, j}/\sum_{k=1}^{|D_i|}\mathbf{d}_{i, j}^{(k)}$.
\end{itemize} 

After prepocessing, we convert $\mathbf{T}$ with $n_c+n_d$ columns to $v_{1:n_c, j}, u_{1:n_c, j}, \mathbf{d}_{1:n_d, j}$. The sum of the dimensions of these vectors is $n_c(m+1) + \sum_{i=1}^{n_d} |D_i|$. This vector is the output of the generator and the input of the discriminator in GAN. Note that GAN does not have access to GMM parameters like $\eta$ and $\sigma$. 

GAN generates $v_{1:n_c, j}, u_{1:n_c, j}, \mathbf{d}_{1:n_d, j}$. Post-processing is straightforward. For continuous variables, we reconstruct $c_{i, j}$ from $u_{i, j}, v_{i, j}$ as $c_{i,j} = 2v_{i, j}\sigma_i^{(k)} + \eta_i^{(k)}$, where $k=\arg_k\max u_{i,j}^{(k)}$. For categorical features, we simply pick the most probable category as $d_{i,j}\leftarrow\argmax_k \mathbf{d}_{i,j}^{(k)}$.

\subsection{Model and data generation}

\begin{figure}[t]
    \centering
    \includegraphics[width=0.7\columnwidth]{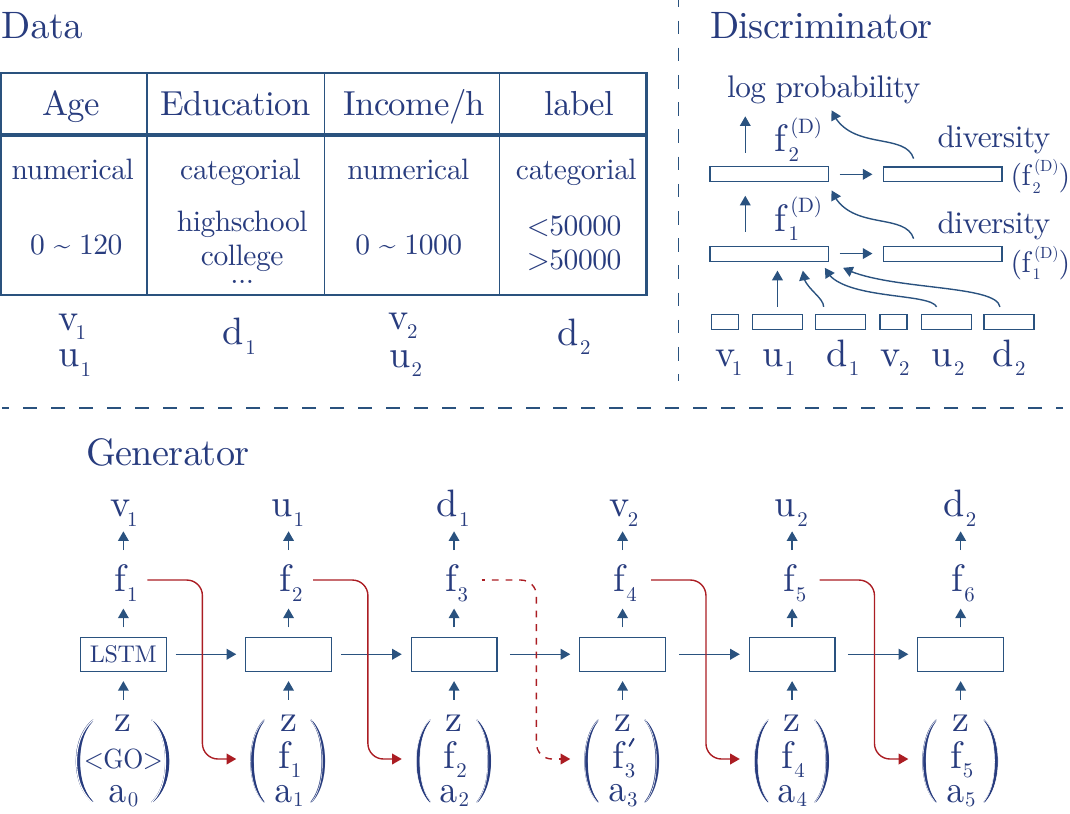}
    \caption{Example of using \tgan to generate a simple census table. The toy example has 2 continuous variables and 2 discrete variables. Our model generates these 4 variables one by one following their original order in the table. Each sample is generated in 6 steps. Each numerical variable is generated in 2 steps while each categorical variable is generated in 1 step. The discriminator concatenates all features together and uses Multi-Layer Perceptron (MLP) to distinguish real and fake data. }
    \label{fig:tgan}
\end{figure}

In GAN, the discriminator $D$ tries to distinguish whether the data is from the real distribution, while the generator $G$ generates synthetic data and tries to fool the discriminator. Figure~\ref{fig:tgan} shows the structure of our \tgan and how to use it to generate tiny tabular data. We use a long-short-term memory (LSTM) network as the generator and use Multi-Layer Perceptron (MLP) in the discriminator. 

\textbf{Generator}: We generate a numerical variable in 2 steps. We first generate the value scalar $v_{i}$, then generate the cluster vector $u_{i}$. We generate categorical feature in 1 step as a probability distribution over all possible labels. 

The output and hidden state size of LSTM is $n_h$. The input to the LSTM in each step $t$ is the random variable $z$, the previous hidden vector $f_{t-1}$ or an embedding vector $f_{t-1}'$ depending on the type of previous output, and the weighted context vector $a_{t-1}$. The random variable $z$ has $n_z$ dimensions. Each dimension is sampled from $\mathcal{N}(0, 1)$. The attention-based context vector $a_t$ is a weighted average over all the previous LSTM outputs $h_{1:t}$. So $a_t$ is a $n_h$-dimensional vector. We learn a attention weight vector $\alpha_t \in \R^t$ and compute context as
\begin{equation}
    a_t = \sum_{k=1}^{t} \frac{\exp \alpha_{t,k}}{\sum_j \exp \alpha_{t,j}} h_k.
\end{equation}
We set $a_0=\mathbf{0}$.
The output of LSTM is $h_t$ and we project the output to a hidden vector $f_t = \tanh(W_h h_t)$, where $W_h$ is a learned parameter in the network. The size of $f_t$ is $n_f$. We further convert the hidden vector to an output variable. 
\begin{itemize}
    \item If the output is the value part of a continuous variable, we compute the output as $v_i = \tanh (W_tf_t)$. The hidden vector for $t+1$ step is $f_t$.
    \item If the output is the cluster part of a continuous variable, we compute the output as $u_i = \text{softmax}(W_tf_t)$. The feature vector for $t+1$ step is $f_t$.
    \item If the output is a discrete variable, we compute the output as  $\mathbf{d}_i = \text{softmax}(W_tf_t)$. The hidden vector for $t+1$ step is $f_t' = E_i[\arg_k\max \mathbf{d}_i]$, where $E \in \R ^{|D_i| \times n_f}$ is an embedding matrix for discrete variable $D_i$. 
    \item $f_0$ is a special vector \texttt{<GO>} and we learn it during the training. 
\end{itemize}

\textbf{Discriminator} 
We use a $l$-layer fully connected neural network as the discriminator. We concatenate $v_{1:n_c}$, $u_{1:n_c}$ and $\mathbf{d}_{1:n_d}$ together as the input. 

We compute the internal layers as 
\begin{align}
    f_1^{(D)} &= \text{LeakyReLU}(\text{BN}(W_1^{(D)} (v_{1:n_c}\oplus u_{1:n_c} \oplus \mathbf{d}_{1:n_d}))),\\
    f_i^{(D)} &= \text{LeakyReLU}(\text{BN}(W_i^{(D)} (f_{i-1}^{(D)} \oplus \text{diversity}(f_{i-1}^{(D)})))), i=2:l,
\end{align}
where $\oplus$ is the concatenation operation. $\text{diversity}(\cdot)$ is the mini-batch discrimination vector \cite{salimans2016improved}. Each dimension of the diversity vector is the total distance between one sample and all other samples in the mini-batch using some learned distance metric. $\text{BN}(\cdot)$ is batch normalization, and $\text{LeakyReLU}(\cdot)$ is the leaky reflect linear activation function. We further compute the output of discriminator as $W^{(D)} (f_l^{(D)} \oplus \text{diversity}(f_l^{(D)}))$ which is a scalar. 

\textbf{Loss Function}
The model is differentiable, so we train our model using Adam optimizer \cite{goodfellow2016deep}. We optimize the generator so that it can fool the discriminator as much as possible. To warm up the model more efficiently, we jointly optimize the KL divergence of discrete variables and the cluster vector of continuous variables by adding them to the loss function. Adding the KL divergence term can also make the model more stable. We optimize generator as 
\begin{equation}
    \mathcal{L}_G = -\E_{z\sim \mathcal{N}(0, 1)}\log D(G(z)) + \sum_{i=1}^{n_c} \text{KL}(u_i', u_i) + \sum_{i=1}^{n_d} \text{KL}(\mathbf{d}_i', \mathbf{d}_i),
\end{equation}
where $u_i'$ and $\mathbf{d}_i'$ are generated data while $u_i$ and $\mathbf{d}_i$ are real data. 
We optimize the discriminator using conventional cross-entropy loss
\begin{equation}
    \mathcal{L}_D = -\E_{v_{1:n_c}, u_{1:n_c}, \mathbf{d}_{1:n_d}\sim \pr(\mathbf{T})}\log D(v_{1:n_c}, u_{1:n_c}, \mathbf{d}_{1:n_d}) +\E_{z\sim \mathcal{N}(0, 1)}\log D(G(z)). 
\end{equation}

\section{Evaluation Setup}\label{seceva}
In this evaluation, we focus on how well \tgan captures the correlation between variables in the table, and whether data scientists can actually use synthetic data to directly learn models. Synthetic data can benefit data science by enabling data scientists to directly learn models over the synthetic data.

\begin{figure}[htb]
    \centering
    \includegraphics[width=0.6\columnwidth]{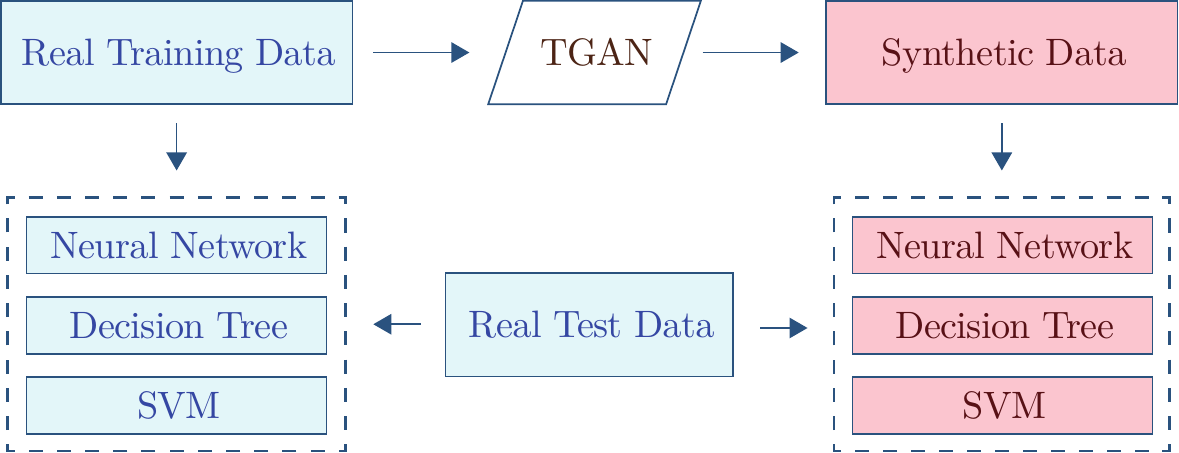}
    \caption{The process of training and evaluating \tgan. The real training data, including the \textit{labels}, is used to learn a GAN and generate synthetic data. Several machine learning models (We choose 5 methods.) are learned using the real training data and the synthetic data. The learned models' accuracy is tested on the real test data that was set aside.}
    \label{fig:eva}
\end{figure}

\noindent \textbf{Machine learning efficacy}:  We first train a \tgan data synthesizer using the real training data $\mathbf{T}$ and generate a synthetic training dataset $\mathbf{T}_{synth}$. We then train machine learning models on both the real and synthetic datasets. We use these trained models on real test data and see how well they perform. Figure~\ref{fig:eva} shows this process of training and evaluating \tgan.

\noindent \textbf{Does it preserve correlation?}: We quantitatively evaluate \tgan's ability to capture correlations between columns by computing the \textit{pairwise} mutual information. We discretize each numeric variable into $20$ buckets. We adjust the boundaries of the buckets so that each bucket has around $5\%$ data. We compute the normalized mutual information as 
\begin{equation}
    \text{NMI}(X, Y) = \frac{1}{\max_{x\in X}\text{E}(x) \times \max_{y\in Y}\text{E}(y)}\sum_{x\in X}\sum_{y\in Y} \pr(x, y) \log \frac{\pr(x, y)}{\pr(x)\pr(y)},
\end{equation}
where $\text{E}(\cdot)$ computes the entropy, and $X, Y$ are two selected columns.

\noindent \textbf{Other data synthesizers}: We compare our \tgan synthesizer with three published methods for generating synthetic data.
\begin{itemize}
    \item \sdv \cite{patki2016synthetic} uses a Gaussian Copula to hierarchically generate multiple synthetic tables in a database.
    \item \dsid \cite{ping2017datasynthesizer} treats each column independently and learns a Bayesian Network for each column. 
    \item \dsco \cite{ping2017datasynthesizer} uses a Bayes Network to model the correlation between columns, and then samples the data from the learned network.
\end{itemize}

\section{Results}\label{secres}
\noindent \textbf{Datasets}: We selected $3$ tabular datasets from the UCI Machine Learning Repository\cite{uci}: (1) The Census-Income dataset, which predicts whether the annual income of a person is above or below \$50k. (2) The KDD Cup 1999 dataset, which predicts the type of malicious internet traffic encountered. (3) The Covertype dataset, \cite{blackard1999comparative} which predicts the forest cover type. Table~\ref{tab:dataset} shows the statistics of the 3 datasets. 

\begin{table}[htb]
\centering
\caption{Dataset statistics. \#Features does not count the label column. \#Labels is number of the unique labels for classification. \#M, \#C, \#D represent Multi-modal Continuous Feature, Continuous Feature, and Discrete Feature respectively. }
\label{tab:dataset}
\small
\begin{tabular*}{\linewidth}{@{\extracolsep{\fill}}lcccccc}
\toprule
\textbf{\normalsize Dataset} & \textbf{\normalsize \#Train} & \textbf{\normalsize \#Test} & \textbf{\normalsize \#Features} & \textbf{\normalsize \#Labels} & \textbf{\normalsize \#M/\#C} & \textbf{\normalsize \#D} \\\midrule
Census           & 199522            & 99761            & 40                   & 2                  & 4/7                                & 33                   \\
KDD99            & 4898431           & 292300           & 41                   & 23                 & 22/27                              & 14                   \\
Covertype        & 465589            & 115423           & 54                   & 7                  & 1/10                               & 44                  \\\bottomrule
\end{tabular*}
\end{table}

\noindent \textbf{How accurate are machine learning models learned on synthetic data?}
To evaluate the machine learning efficacy of the synthetic data, we follow the process described in Figure~\ref{fig:eva}.   

The Census dataset is imbalanced. About $93\%$ of the samples have positive labels, while only $7\%$ have negative labels. Because simply predicting the majority can achieve significantly high accuracy, we used Macro-F1 to evaluate the performance. We selected $5$ well-known models: \texttt{Decision Tree}, \texttt{Linear Support Vector Machine}, \texttt{Random Forest}, \texttt{AdaBoost}, and \texttt{Multi-Layer Perceptron}.

\begin{table}[t]
\centering
\caption{Macro-F1 evaluation of machine learning models trained on Census dataset. `-' indicates the machine learning model deteriorates into predicting the majority. }
\label{tab:census-res}
\small
\begin{tabular*}{\linewidth}{@{\extracolsep{\fill}}rlllll}
\toprule
\textbf{\normalsize Method}          & \textbf{\normalsize Real} & \textbf{\normalsize \sdv}   & \textbf{\normalsize \dsid} & \textbf{\normalsize \dsco} & \textbf{\normalsize \tgan}  \\\midrule
DT & & & & & \\
$\mathtt{max\_depth}=10$        & 74.65        & 48.61 & 32.26  & 32.24   & \hl{68.70} \\
$\mathtt{max\_depth}=20$        & 75.11        & 48.64 & 31.16  & 31.77   & \hl{64.42} \\[1.5ex]
SVM             & 71.30        & -     & -      & 25.69   & \hl{67.77} \\[1.5ex]
RF & & & & & \\
$\mathtt{max\_depth}=10$, $\mathtt{estimators=10}$ & 59.04        & -     & -      & -       & \hl{51.42} \\
$\mathtt{max\_depth}=20$, $\mathtt{estimators=10}$ & 70.95        & -     & -      & 32.26   & \hl{65.89} \\[1.5ex]
AdaBoost        & 74.10        & -     & -      & 32.27   & \hl{70.08} \\[1.5ex]
MLP & & & & & \\
$\mathtt{layer\_sizes}=(100,)$         & 75.47        & 53.15 & 25.5   & 26.34   & \hl{71.81} \\
$\mathtt{layer\_sizes}=(200, 200)$    & 73.94        & -     & -      & 32.14   & \hl{68.75}
\\\bottomrule
\end{tabular*}
\end{table}

\begin{table}[t]
\centering
\caption{Accuracy of machine learning models trained on the real and synthetic training set. (\dsco fails on KDD99 dataset.)}
\label{tab:other-res}
\small
\begin{tabular*}{\linewidth}{@{\extracolsep{\fill}}rlll llll}
\toprule
 & \multicolumn{3}{c}{\textbf{\normalsize KDD99}} & \multicolumn{4}{c}{\textbf{\normalsize covertype}} \\[0.5ex]
             \cline{2-4} \cline{5-8}\\[-1.5ex]
\textbf{\normalsize Model}        & \textbf{\normalsize Real}    & \textbf{\normalsize \sdv}    & \textbf{\normalsize \tgan}    & \textbf{\normalsize Real}     & \textbf{\normalsize \sdv}      & \textbf{\normalsize \dsco}      & \textbf{\normalsize \tgan}     \\[0.5ex]
\midrule\\[-1.5ex]
DT & & & & & & \\
$\mathtt{max\_depth}=10$    & 97.75   & 58.34  & \hl{90.14}  & 77.43    & 46.10   & 48.76  & \hl{69.27}   \\
$\mathtt{max\_depth}=30$    & 97.35   & 56.46  & \hl{80.58}  & 90.82    & 36.83   & 46.21 & \hl{58.88}   \\[1.5ex]
SVM          & 93.64   & 56.15  & \hl{94.56}  & 70.97    & 46.30    &48.76 & \hl{67.94}   \\[1.5ex]
RF & & & & & & \\
$\mathtt{max\_depth}=10$, $\mathtt{estimators=10}$ & 97.79   & 60.61  & \hl{93.36}  & 74.58    & 45.30   & 48.91  & \hl{66.60}    \\
$\mathtt{max\_depth}=20$, $\mathtt{estimators=10}$ & 97.81   & 56.46  & \hl{92.33}  & 85.13    & 46.78   &48.85 & \hl{69.33}   \\[1.5ex]
AdaBoost     & 19.93   & \hl{75.94}  & 40.43  & 49.81    & 40.95   & 48.88 & \hl{66.11}   \\[1.5ex]
MLP & & & & & & \\
$\mathtt{layer\_sizes}=(100,)$      & 97.48   & 56.15  & \hl{95.91}  & 60.32    & 47.82   &48.86 & \hl{61.24}   \\
$\mathtt{layer\_sizes}=(200, 200)$ & 96.08   & 56.14  & \hl{66.38}  & 84.11    & 46.97   & 48.79 & \hl{68.80}   \\[0.5ex]\bottomrule
\end{tabular*}
\end{table}

\begin{table}[b]
\centering
\caption{Distance between NMI matrices of real data and synthetic data generated by different methods. }
\label{tab:mutual}
\small
\begin{tabular*}{\linewidth}{@{\extracolsep{\fill}}l lll lll}
\toprule
          & \multicolumn{3}{c}{\textbf{\normalsize RMSE}}                    & \multicolumn{3}{c}{\textbf{\normalsize MAE}}                       \\
          \cmidrule{2-4}\cmidrule{5-7}
\textbf{\normalsize Dataset}   & \textbf{\normalsize \sdv}          & \textbf{\normalsize \dsco}       & \textbf{\normalsize GAN}           & \textbf{\normalsize \sdv}           & \textbf{\normalsize \dsco}        & \textbf{\normalsize GAN}           \\\midrule
Census    & 0.1474 & 0.1731 & \hl{0.0673} & 0.0657 & 0.0696 & \hl{0.0295} \\
KDD99     & 0.4902 & -            & \hl{0.4551}  & 0.2753  & -             & \hl{0.2476}  \\
Covertype & 0.4751 & 0.1252 & \hl{0.1185} & 0.2852 & 0.0256 & \hl{0.0192} \\\bottomrule            
\end{tabular*}
\end{table}

Table~\ref{tab:census-res} shows the Macro-F1 on the Census dataset. We observe that although the Census dataset is simple in a prediction sense, it is challenging in terms of synthetic data generation. 
\begin{itemize}
    \item \sdv fails to capture the internal relations between columns, so all machine learning models can do nothing but predict the majority.
    \item \dsid does not consider the relations between features, so it also fails in training a machine learning model. 
    \item \dsco is extremely time-consuming in that it is impossible to train it with all the real data. We trained it with $50000$ samples, and the maximum degree in the Bayes Network is $2$.
    \item \tgan performs reasonably well. The average performance gap between real data and synthetic data is $5.7\%$ comparing with $24.9\%$ for \sdv and $43.3\%$ for \dsco. 
    \item \tgan data in many cases keep the ranking of different machine learning models. This is important because it indicates that a data scientist can evaluate machine learning models on the synthetic dataset and select the best model. 
\end{itemize}
For the KDD99, and Covertype datasets, we use accuracy to compare different methods. Table~\ref{tab:other-res} shows the accuracy of the KDD99 and Covertype datasets. \tgan also consistently outperforms other data synthesizers. 

\begin{figure}[t]
    \centering
    \adjustbox{valign=m}{%
\begin{tikzpicture}[font=\sffamily\sansmath] 
\begin{axis}[width=4.4cm,height=4.4cm,
xmin=1,xmax=41,ymin=1,ymax=41,
colormap name=blueblue,
ticks=none
]
\addplot[matrix plot,point meta=explicit] 
table[x index=0,y index=1,meta index=2] {csv/census-MI-real.csv.dat};
\end{axis}
\end{tikzpicture}
}%
\adjustbox{valign=m}{%
\begin{tikzpicture}[font=\sffamily\sansmath] 
\begin{axis}[width=4.4cm,height=4.4cm,
xmin=1,xmax=41,ymin=1,ymax=41,
colormap name=blueblue,
ticks=none
]
\addplot[matrix plot,point meta=explicit] 
table[x index=0,y index=1,meta index=2] {csv/census-MI-sdv.csv.dat};
\end{axis}
\end{tikzpicture}
}%
\adjustbox{valign=m}{%
\begin{tikzpicture}[font=\sffamily\sansmath] 
\begin{axis}[width=4.4cm,height=4.4cm,
xmin=1,xmax=41,ymin=1,ymax=41,
colormap name=blueblue,
ticks=none
]
\addplot[matrix plot,point meta=explicit] 
table[x index=0,y index=1,meta index=2] {csv/census-MI-ds.csv.dat};
\end{axis}
\end{tikzpicture}
}%
\adjustbox{valign=m}{%
\begin{tikzpicture}[font=\sffamily\sansmath] 
\begin{axis}[width=4.4cm,height=4.4cm,
xmin=1,xmax=41,ymin=1,ymax=41,
colormap name=blueblue,
colorbar,
ticks=none
]
\addplot[matrix plot,point meta=explicit] 
table[x index=0,y index=1,meta index=2] {csv/census-MI-gan.csv.dat};
\end{axis}
\end{tikzpicture}
}%
    \caption{Inspect NMI matrices for Census dataset. From left to right: real data, \sdv, \dsco and \tgan. }
    \label{fig:census-mutual}
\smallskip
    \adjustbox{valign=m}{%
\begin{tikzpicture}[font=\sffamily\sansmath] 
\begin{axis}[width=4.4cm,height=4.4cm,
xmin=1,xmax=42,ymin=1,ymax=42,
colormap name=blueblue,
ticks=none
]
\addplot[matrix plot,point meta=explicit] 
table[x index=0,y index=1,meta index=2] {csv/kdd-MI-real.csv.dat};
\end{axis}
\end{tikzpicture}
}%
\adjustbox{valign=m}{%
\begin{tikzpicture}[font=\sffamily\sansmath] 
\begin{axis}[width=4.4cm,height=4.4cm,
xmin=1,xmax=42,ymin=1,ymax=42,
colormap name=blueblue,
ticks=none
]
\addplot[matrix plot,point meta=explicit] 
table[x index=0,y index=1,meta index=2] {csv/kdd-MI-gan.csv.dat};
\end{axis}
\end{tikzpicture}
}%
\adjustbox{valign=m}{%
\begin{tikzpicture}[font=\sffamily\sansmath] 
\begin{axis}[width=4.4cm,height=4.4cm,
xmin=1,xmax=55,ymin=1,ymax=55,
colormap name=blueblue,
ticks=none
]
\addplot[matrix plot,point meta=explicit] 
table[x index=0,y index=1,meta index=2] {csv/cover-MI-real.csv.dat};
\end{axis}
\end{tikzpicture}
}%
\adjustbox{valign=m}{%
\begin{tikzpicture}[font=\sffamily\sansmath] 
\begin{axis}[width=4.4cm,height=4.4cm,
xmin=1,xmax=55,ymin=1,ymax=55,
colormap name=blueblue,
colorbar,
ticks=none
]
\addplot[matrix plot,point meta=explicit] 
table[x index=0,y index=1,meta index=2] {csv/cover-MI-gan.csv.dat};
\end{axis}
\end{tikzpicture}
}%
    \caption{Inspect NMI matrices for KDD99 and Covertype dataset. From left to right: real data (KDD99), \tgan (KDD99), real data (covertype), \tgan (covertype). }
    \label{fig:mutual}
\smallskip
    \scalebox{0.45}{%
\begin{tikzpicture}[font=\sffamily\sansmath] 
\begin{axis}[width=8cm,height=6cm,
xmin=0,xmax=20,
xticklabel style={font=\sffamily\sansmath\normalsize},
yticklabel style={font=\sffamily\sansmath\normalsize},
yticklabel style={/pgf/number format/.cd,fixed,fixed zerofill,precision=0,1000 sep={}}, 
ymin=0,ymax=300,
x label style={font=\sffamily\Large},y label style={font=\sffamily\Large},
tick align = outside,xtick pos=left,ytick pos=left
]
\addplot[ybar interval,fill=blue,draw=blue] table[x index = 0,y index = 1, col sep=comma] {csv/census-nn-real.csv};
\end{axis}
\end{tikzpicture}
}%
\scalebox{0.45}{%
\begin{tikzpicture}[font=\sffamily\sansmath] 
\begin{axis}[width=8cm,height=6cm,
xmin=0,xmax=100,
xticklabel style={font=\sffamily\sansmath\normalsize},
yticklabel style={font=\sffamily\sansmath\normalsize},
yticklabel style={/pgf/number format/.cd,fixed,fixed zerofill,precision=0,1000 sep={}}, 
ymin=0,ymax=100,
x label style={font=\sffamily\Large},y label style={font=\sffamily\Large},
tick align = outside,xtick pos=left,ytick pos=left
]
\addplot[ybar interval,fill=blue,draw=blue] table[x index = 0,y index = 1, col sep=comma] {csv/census-nn-sdv.csv};
\end{axis}
\end{tikzpicture}
}%
\scalebox{0.45}{%
\begin{tikzpicture}[font=\sffamily\sansmath] 
\begin{axis}[width=8cm,height=6cm,
xmin=0,xmax=100,
xticklabel style={font=\sffamily\sansmath\normalsize},
yticklabel style={font=\sffamily\sansmath\normalsize},
yticklabel style={/pgf/number format/.cd,fixed,fixed zerofill,precision=0,1000 sep={}}, 
ymin=0,ymax=1000,
x label style={font=\sffamily\Large},y label style={font=\sffamily\Large},
tick align = outside,xtick pos=left,ytick pos=left
]
\addplot[ybar interval,fill=blue,draw=blue] table[x index = 0,y index = 1, col sep=comma] {csv/census-nn-ds.csv};
\end{axis}
\end{tikzpicture}
}%
\scalebox{0.45}{%
\begin{tikzpicture}[font=\sffamily\sansmath] 
\begin{axis}[width=8cm,height=6cm,
xmin=0,xmax=20,
xticklabel style={font=\sffamily\sansmath\normalsize},
yticklabel style={font=\sffamily\sansmath\normalsize},
yticklabel style={/pgf/number format/.cd,fixed,fixed zerofill,precision=0,1000 sep={}}, 
ymin=0,ymax=300,
x label style={font=\sffamily\Large},y label style={font=\sffamily\Large},
tick align = outside,xtick pos=left,ytick pos=left
]
\addplot[ybar interval,fill=blue,draw=blue] table[x index = 0,y index = 1, col sep=comma] {csv/census-nn-gan.csv};
\end{axis}
\end{tikzpicture}
}%
    \captionof{figure}{\label{fig:census-nn}The distribution of the distance to the nearest neighbor on Census dataset. From left to right:  $\mathbf{T}_{real}', \mathbf{T}_{GC}', \mathbf{T}_{BN\_Co}', \mathbf{T}_{TGAN}'$} 
\smallskip
    \scalebox{0.45}{%
\begin{tikzpicture}[font=\sffamily\sansmath] 
\begin{axis}[width=8cm,height=6cm,
xmin=0,xmax=5,
xticklabel style={font=\sffamily\sansmath\normalsize},
yticklabel style={font=\sffamily\sansmath\normalsize},
yticklabel style={/pgf/number format/.cd,fixed,fixed zerofill,precision=0,1000 sep={}}, 
ymin=0,ymax=1000,
x label style={font=\sffamily\Large},y label style={font=\sffamily\Large},
tick align = outside,xtick pos=left,ytick pos=left
]
\addplot[ybar interval,fill=blue,draw=blue] table[x index = 0,y index = 1, col sep=comma] {csv/kdd-nn-real.csv};
\end{axis}
\end{tikzpicture}
}%
\scalebox{0.45}{%
\begin{tikzpicture}[font=\sffamily\sansmath] 
\begin{axis}[width=8cm,height=6cm,
xmin=0,xmax=5,
xticklabel style={font=\sffamily\sansmath\normalsize},
yticklabel style={font=\sffamily\sansmath\normalsize},
yticklabel style={/pgf/number format/.cd,fixed,fixed zerofill,precision=0,1000 sep={}}, 
ymin=0,ymax=1000,
x label style={font=\sffamily\Large},y label style={font=\sffamily\Large},
tick align = outside,xtick pos=left,ytick pos=left
]
\addplot[ybar interval,fill=blue,draw=blue] table[x index = 0,y index = 1, col sep=comma] {csv/kdd-nn-gan.csv};
\end{axis}
\end{tikzpicture}
}%
\scalebox{0.45}{%
\begin{tikzpicture}[font=\sffamily\sansmath] 
\begin{axis}[width=8cm,height=6cm,
xmin=0,xmax=10,
xticklabel style={font=\sffamily\sansmath\normalsize},
yticklabel style={font=\sffamily\sansmath\normalsize},
yticklabel style={/pgf/number format/.cd,fixed,fixed zerofill,precision=0,1000 sep={}}, 
ymin=0,ymax=250,
x label style={font=\sffamily\Large},y label style={font=\sffamily\Large},
tick align = outside,xtick pos=left,ytick pos=left
]
\addplot[ybar interval,fill=blue,draw=blue] table[x index = 0,y index = 1, col sep=comma] {csv/cover-nn-real.csv};
\end{axis}
\end{tikzpicture}
}%
\scalebox{0.45}{%
\begin{tikzpicture}[font=\sffamily\sansmath] 
\begin{axis}[width=8cm,height=6cm,
xmin=0,xmax=10,
xticklabel style={font=\sffamily\sansmath\normalsize},
yticklabel style={font=\sffamily\sansmath\normalsize},
yticklabel style={/pgf/number format/.cd,fixed,fixed zerofill,precision=0,1000 sep={}}, 
ymin=0,ymax=250,
x label style={font=\sffamily\Large},y label style={font=\sffamily\Large},
tick align = outside,xtick pos=left,ytick pos=left
]
\addplot[ybar interval,fill=blue,draw=blue] table[x index = 0,y index = 1, col sep=comma] {csv/cover-nn-gan.csv};
\end{axis}
\end{tikzpicture}
}%
    \captionof{figure}{\label{fig:nn}The distribution of the distance to the nearest neighbor on KDD99 and Covertype dataset. From left to right:  $\mathbf{T}_{real}'$(KDD99), $\mathbf{T}_{TGAN}'$(KDD99), $\mathbf{T}_{real}'$(Covertype), $\mathbf{T}_{TGAN}'$(Covertype)}
\end{figure}

\noindent \textbf{Are correlations between variables preserved?}
 Figure~\ref{fig:census-mutual} visualizes the NMI matrices of real data, \sdv, \dsco and \tgan on the Census dataset. Figure~\ref{fig:mutual} visualizes the NMI matrices of real data and \tgan on the KDD99 and Covertype datasets. When compared with \sdv and \dsco, \tgan learns the correlation between variables significantly better. Table~\ref{tab:mutual} quantitatively shows the root mean square error and mean absolute error between NMI matrices of real data and synthetic data generated by different methods. 

\noindent \textbf{How close is it to real data?} 
 Next, we attempt to answer the question: \textit{``Is the data synthesizer simply remembering the data in the training set?"}. To assess this quantitatively we follow these steps. We sample $10000$ data points from the training set as $\mathbf{T}_{standard}$. We sample another $1000$ data points from the test set $\mathbf{T}_{real}'$. We also sample $1000$ data points from different \sdv, \dsco, and \tgan synthetic data as $\mathbf{T}_{GC}', \mathbf{T}_{BN\_Co}', \mathbf{T}_{TGAN}'$. For data in $\mathbf{T}'$, we compute the distance between $\mathbf{T}'$ and $\mathbf{T}_{standard}$. We compute the distances between $c_{1:n_c, j}, \mathbf{d}_{1:n_d, j}$ from $\mathbf{T}_{standard}$ and $c'_{1:n_c, k}, \mathbf{d}'_{1:n_d, k}$ from $\mathbf{T}'$ as 
\begin{equation}
    distance(c_{1:n_c,j}, \mathbf{d}_{1:n_d,j}, c_{1:n_c,k}', \mathbf{d}_{1:n_d,k}') = \sum_{i=1}^{n_c} \frac{1}{\text{std}(c_{i})}|c_{i, j} - c'_{i,k}| + \sum_{i=1}^{n_d} \text{neq}(\mathbf{d}_{i,j}, \mathbf{d}'_{i,k}).
\end{equation}
Where $\text{neq}(x, y)$ is $1$ if $x$ and $y$ are different, and $0$ otherwise. Figures~\ref{fig:census-nn} and ~\ref{fig:nn} show the histogram of nearest neighbor distances on three datasets. The nearest neighbor distance distribution of \tgan is very close to real data.

\section{Conclusion}\label{seccon}
In this paper, we propose \tgan, a GAN-based model for generating relational tables containing continuous and discrete variables. We cluster numerical variables to deal with the multi-modal distribution for continuous features. We add noise and KL divergence into the loss function to effectively generate discrete features. We observe that GANs can effectively capture the correlations between features and are more scalable for large datasets. We show that our model can generate high-quality synthetic data to benefit data science. Relational databases are widely used and very difficult to model. Our model only supports a single table with numerical and categorical features. In the future, we would explore how to model sequential data and how to model multiple tables using GAN. 

\bibliographystyle{plainnat}
\bibliography{nips_2018}

\end{document}